\documentclass{article}
\usepackage{ijcai23}

\usepackage{times}
\usepackage{soul}
\usepackage{url}
\usepackage[hidelinks]{hyperref}
\usepackage[utf8]{inputenc}
\usepackage[small]{caption}
\usepackage{graphicx}
\usepackage{amsmath}
\usepackage{amsthm}
\usepackage{booktabs}
\usepackage{algorithm}
\usepackage[switch]{lineno}

\usepackage{amsfonts, amssymb}
\usepackage{multirow}
\usepackage{subcaption}

\usepackage{pifont}
\newcommand{\cmark}{\ding{51}}%
\newcommand{\xmark}{\ding{55}}%

\usepackage[normalem]{ulem}
\useunder{\uline}{\ul}{}

\usepackage{algpseudocode}
\algrenewcommand\algorithmicindent{0.9em}
\makeatletter
\let\OldStatex\Statex
\renewcommand{\Statex}[1][3]{%
  \setlength\@tempdima{\algorithmicindent}%
  \OldStatex\hskip\dimexpr#1\@tempdima\relax}
\makeatother

\newtheorem{definition}{Definition}

\title{FedHGN: A Federated Framework for Heterogeneous Graph Neural Networks}

\author{
Xinyu Fu
\and
Irwin King
\affiliations
The Chinese University of Hong Kong, Hong Kong, China
\emails
\{xyfu, king\}@cse.cuhk.edu.hk
}

\begin{document}

\maketitle

\begin{abstract}
Heterogeneous graph neural networks (HGNNs) can learn from typed and relational graph data more effectively than conventional GNNs.
With larger parameter spaces, HGNNs may require more training data, which is often scarce in real-world applications due to privacy regulations (e.g., GDPR).
Federated graph learning (FGL) enables multiple clients to train a GNN collaboratively without sharing their local data.
However, existing FGL methods mainly focus on homogeneous GNNs or knowledge graph embeddings; few have considered heterogeneous graphs and HGNNs.
In federated heterogeneous graph learning, clients may have private graph schemas. Conventional FL/FGL methods attempting to define a global HGNN model would violate \emph{schema privacy}.
To address these challenges, we propose FedHGN, a novel and general FGL framework for HGNNs.
FedHGN adopts \emph{schema-weight decoupling} to enable schema-agnostic knowledge sharing and employs \emph{coefficients alignment} to stabilize the training process and improve HGNN performance.
With better privacy preservation, FedHGN consistently outperforms local training and conventional FL methods on three widely adopted heterogeneous graph datasets with varying client numbers.
The code is available at \url{https://github.com/cynricfu/FedHGN}.
\end{abstract}

\section{Introduction}
Graph neural networks (GNNs)~\cite{DBLP:conf/nips/DefferrardBV16,DBLP:conf/iclr/KipfW17,DBLP:conf/nips/HamiltonYL17,DBLP:conf/iclr/VelickovicCCRLB18,DBLP:conf/www/ZhangZMKK22,DBLP:journals/corr/abs-2212-01026} combine graph representation learning and deep learning to handle graph-structured data more effectively than traditional methods~\cite{DBLP:conf/kdd/PerozziAS14,DBLP:conf/www/TangQWZYM15,DBLP:conf/kdd/GroverL16,DBLP:conf/kdd/ZhangZSKK22}. GNNs have various applications, such as community detection~\cite{DBLP:conf/iclr/ChenLB19,DBLP:conf/aaai/JinLLHZ19}, recommender systems~\cite{DBLP:journals/corr/BergKW17,DBLP:conf/kdd/YingHCEHL18,DBLP:conf/www/000100LK22,DBLP:conf/wsdm/00010ZZMHK22}, traffic forecasting~\cite{DBLP:conf/iclr/LiYS018,DBLP:conf/uai/ZhangSXMKY18,DBLP:conf/ijcai/YuYZ18}, and text similarity~\cite{ma2022graph}.
However, many real-world graphs are \emph{heterogeneous graphs}, which have multiple types of nodes and edges. For instance, a financial network may consist of customers, merchants, transactions, and various relationships among them. Therefore, researchers have developed heterogeneous GNNs (HGNNs), such as RGCN~\cite{DBLP:conf/esws/SchlichtkrullKB18}, to capture the complex semantics in heterogeneous graph structures.


Similar to other deep learning models, GNN performance depends on the size of the training data. If the training data is insufficient, GNNs may overfit and fail to generalize to unseen data. This problem is more severe for HGNNs, which have much larger parameter spaces~\cite{DBLP:conf/esws/SchlichtkrullKB18}. Moreover, in reality, adequate training data is often unavailable, which hinders HGNN's performance.





One possible solution to data scarcity is to collect and integrate samples from multiple parties to create a large shared dataset for model training. However, this approach raises serious privacy concerns in many application scenarios, which prevent HGNNs from using training data in a centralized way. For instance, banks and insurance companies may want to collaborate on developing an HGNN for better fraud detection. But their data, such as customer information and transaction records, are highly sensitive. They cannot transmit these data to another place for HGNN training. Therefore, there is a trade-off between data availability and data privacy. HGNNs trained separately at each client (e.g., a financial institution in this case) would perform poorly due to limited training data, while centralized training by aggregating data from all clients is not feasible due to commercial or regulatory reasons (e.g., GDPR). This leads to a question: can we train an HGNN that leverages each client's training data without violating local data privacy?

Federated learning (FL) is a promising technique to achieve this goal. Instead of sharing private data, FL clients upload model weights to a central server which aggregates them to improve the global model.
The collaboratively trained model is expected to outperform the models trained locally at each client.
Previous FL frameworks mainly focused on computer vision (CV) and natural language processing (NLP) tasks~\cite{DBLP:conf/aistats/McMahanMRHA17,DBLP:conf/mlsys/LiSZSTS20,DBLP:conf/aaai/0007LGKL22}.
Federated graph learning (FGL), i.e., FL on graph-structured data, has just emerged in recent years.
Some works have explored FGL in homogeneous graphs~\cite{DBLP:conf/nips/ZhangYLSY21,DBLP:conf/nips/XieMXY21}, recommender systems~\cite{DBLP:journals/corr/abs-2102-04925}, and knowledge graphs~\cite{DBLP:conf/jist/ChenZYJC21,DBLP:conf/cikm/PengLSZ021}.
However, FL with heterogeneous graphs and HGNNs, which have a great need for collaborative training, has received little attention.

\begin{figure}[t]
\centering
\includegraphics[width=0.75\columnwidth]{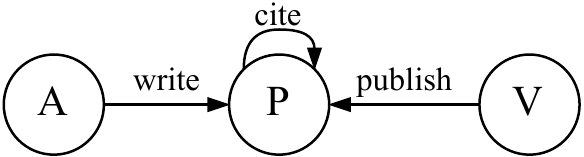}
\caption{An example graph schema of a scholar network consisting of authors (A), papers (P), and venues (V).}
\label{fig:schema}
\end{figure}

Real-world federated heterogeneous graph learning faces two unique challenges. First, unlike homogeneous graphs, heterogeneous graphs have special metadata called graph schemas. These are meta templates that describe the node types and edge types of the graph. For example, Figure~\ref{fig:schema} shows a graph schema of a scholar network. The problem is , \emph{clients may not want to share the schema information}, which is essential to define an HGNN model. The graph schema reveals high-level information about how the graph is constructed, which may be sensitive in some domains, such as pharmaceutical companies and financial institutions.



Second, even if schema sharing is allowed among clients, \emph{it is usually difficult to match node/edge types of different graphs with each other}, which is crucial for HGNNs to reduce complexity and share knowledge. In a federated system, different clients may use different pipelines to construct their heterogeneous graphs. For instance, one client may label their paper nodes as \textit{paper} while another may call them \textit{article}. This matching process may not always have the domain knowledge and human expertise available. Moreover, when a new client joins the federated system, this schema matching process needs to be repeated, and the HGNN may need to be re-defined to include new node/edge types, both of which are undesirable for industrial FL applications.


Motivated by the lack of an FGL framework for HGNNs and the challenges above, we propose \textbf{FedHGN}, a \emph{\underline{Fed}erated learning framework for \underline{HGN}Ns} that preserves schema privacy and avoids schema matching. The key idea of FedHGN is to eliminate the direct correspondence between graph schemas and HGNN model weights. We achieve this by applying \emph{schema-weight decoupling} to decompose HGNN weights into shareable schema-agnostic basis weights and confidential schema-specific coefficients. To prevent the potential discrepancy between coefficients of the same/similar schemas at different clients, FedHGN employs \emph{coefficients alignment} to minimize the distance between coefficients of most similar node/edge types. Thus, FedHGN enables collaborative training for better-performing HGNNs without sacrificing schema privacy or necessitating schema matching.


In summary, we make the following main contributions:
\begin{itemize}
    \item We propose FedHGN, a novel FGL framework that enables collaborative HGNN training among clients with private heterogeneous graphs. To our knowledge, this is the first work for general federated HGNN learning.
    \item We propose a schema-weight decoupling (SWD) strategy to achieve schema-agnostic knowledge sharing, which preserves schema-level privacy among clients.
    \item We design a schema coefficients alignment (CA) component via a novel regularization term, which stabilizes FedHGN training and improves HGNN performance.
    \item We conduct extensive experiments on benchmark datasets with varying numbers of clients for node classification. Results show that FedHGN consistently outperforms local training and conventional FL methods.
\end{itemize}

\section{Related Work}
In this section, we review recent studies related to this work. Section~\ref{sec:gnn_survey} introduces graph neural networks (GNNs) and especially heterogeneous GNNs (HGNNs). Section~\ref{sec:fl_survey} introduces conventional federated learning (FL) and recent efforts on federated graph learning (FGL).

\subsection{Graph Neural Networks} \label{sec:gnn_survey}

Graph neural networks (GNNs) are designed to overcome the limitations of shallow graph embedding methods, such as DeepWalk~\cite{DBLP:conf/kdd/PerozziAS14} and node2vec~\cite{DBLP:conf/kdd/GroverL16}.
Taking advantage of the neural network architecture, GNNs can leverage node attributes and benefit from various training paradigms of deep learning.
The general idea of GNNs is to characterize each node by its own features and its local neighborhood's information.
Following this idea and combined with graph signal processing, spectral-based GNNs like ChebNet~\cite{DBLP:conf/nips/DefferrardBV16} and GCN~\cite{DBLP:conf/iclr/KipfW17} were first developed to perform convolutions in the Fourier domain of a graph.
Then, by allowing convolutions in the graph domain, spatial-based GNNs like GraphSAGE~\cite{DBLP:conf/nips/HamiltonYL17} and GAT~\cite{DBLP:conf/iclr/VelickovicCCRLB18} were proposed to improve scalability and generalization ability,
with many follow-up studies~\cite{DBLP:conf/iclr/XuHLJ19,DBLP:conf/icml/WuSZFYW19}.

\subsubsection{Heterogeneous GNNs}
Conventional GNNs assume homogeneous input graphs with single node/edge types.
However, real-world graphs are usually heterogeneous, having multiple node/edge types.
To capture complex structural and semantic heterogeneity in these graphs, researchers proposed heterogeneous GNNs (HGNNs).
As an early attempt, RGCN~\cite{DBLP:conf/esws/SchlichtkrullKB18} aggregates neighborhoods using relation-specific weights.
Many other HGNNs follow similar ideas to project and aggregate neighborhoods based on node/edge types, including HetGNN~\cite{DBLP:conf/kdd/ZhangSHSC19}, HGT~\cite{DBLP:conf/www/HuDWS20}, Simple-HGN~\cite{DBLP:conf/kdd/LvDLCFHZJDT21}, etc.
Another line of research considers meta-paths to capture complex high-order relationships between nodes, such as HAN~\cite{DBLP:conf/www/WangJSWYCY19}, MAGNN~\cite{DBLP:conf/www/0004ZMK20}, and GTN~\cite{DBLP:conf/nips/YunJKKK19}.

Although many HGNNs claim to have good performance in real-world applications, none have considered the federated setting with privacy concerns across clients.
They simply assume a centralized setting with all data readily available.
With privacy restrictions, HGNNs trained with limited local data could suffer from degraded performance and biased predictions.
Our proposed FedHGN offers a feasible way to collaboratively train better-performing HGNNs among multiple clients without sharing local graph data and graph schemas.

\subsection{Federated Learning} \label{sec:fl_survey}

Federated learning (FL) is a machine learning setting for collaborative cross-client model training without sharing raw data~\cite{DBLP:journals/corr/abs-2007-06849,DBLP:conf/www/ZhangZLXK23}.
Instead of gathering data to a central server, FL keeps data localized on each client and only shares model weights or gradients.
Multiple FL algorithms have been proposed for regular data in Euclidean space, such as FedAvg~\cite{DBLP:conf/aistats/McMahanMRHA17}, FedProx~\cite{DBLP:conf/mlsys/LiSZSTS20}, and SCAFFOLD~\cite{DBLP:conf/icml/KarimireddyKMRS20}.
These FL algorithms are applied in CV and NLP, where data samples are naturally isolated from each other.
But for graphs, especially in node-level and link-level tasks, nodes and edges as data samples are associated with each other. Naively migrating conventional FL frameworks might not bring optimal results for graph data.

\subsubsection{Federated Graph Learning}
Recently, some researchers have proposed FL methods for graphs, which are coined as federated graph learning (FGL).
GCFL~\cite{DBLP:conf/nips/XieMXY21} employs a client clustering mechanism to reduce data heterogeneity in federated graph classification.
FedSage~\cite{DBLP:journals/corr/abs-2102-04925} generates missing neighbors to mend dropped cross-client links caused by data isolation in an FGL system.
FedGNN~\cite{DBLP:journals/corr/abs-2102-04925} conducts GNN-based recommendations on user devices with a privacy-preserving FGL framework.
FedE~\cite{DBLP:conf/jist/ChenZYJC21} and FKGE~\cite{DBLP:conf/cikm/PengLSZ021} are federated algorithms for knowledge graph embedding (KGE).
FedAlign~\cite{DBLP:journals/corr/abs-2011-11369} applies basis alignment and weight constraint on federated RGCN to improve personalization and convergence.

However, few FGL algorithms have considered heterogeneous graphs and HGNNs.
Aside from graph data (nodes/edges) and optional attributes, heterogeneous graphs also have special metadata called graph schemas.
Naively applying existing FGL methods would lead to schema privacy breaches or require schema matching.
FedE, FKGE, and FedAlign are the most related research works to our FedHGN, but they still have some notable limitations.
Due to the inherent restriction of KGE algorithms, FedE and FKGE assume all or part of the entities/relations are already matched among clients. While our FedHGN does not require any entity/relation matching, reducing privacy concerns and human labor.
FedAlign focuses on a particular HGNN model, namely RGCN. On the other hand, our FedHGN is a general framework compatible with almost any HGNN architecture. Moreover, FedHGN is equipped with the coefficients alignment component that can alleviate the discrepancy between same-type schema coefficients at different clients.

\section{Preliminary}

This section gives formal definitions of some important graph and FL terminologies related to this work.
Table~\ref{tab:notation} provides a quick reference to the notations used in this work.

\begin{definition}[Heterogeneous Graph]
A heterogeneous graph is defined as a graph $\mathcal{G}=\left(\mathcal{V},\mathcal{E}\right)$ associated with a node type mapping function $\phi : \mathcal{V} \rightarrow \mathcal{A}$ and an edge type mapping function $\psi : \mathcal{E} \rightarrow \mathcal{R}$.
$\mathcal{A}$ and $\mathcal{R}$ denote the pre-defined sets of node types and edge types, respectively, with $|\mathcal{A}|+|\mathcal{R}|>2$.
The graph schema of $\mathcal{G}$ is given by $\mathcal{T}_\mathcal{G}=(\mathcal{A}, \mathcal{R})$, which is essentially a graph of node types connected by edge types.
\end{definition}

\begin{definition}[Heterogeneous Graph Neural Network]
\label{def:hgnn}
A heterogeneous graph neural network (HGNN) is a GNN model designed for representation learning on heterogeneous graphs. Generally, an HGNN layer can be formulated as
\begin{equation}
    \mathbf{h}_v = \operatorname{Trans}_{\phi(v)}\left(\operatorname{Reduce}\left(\left\{\operatorname{Agg}_{r}(\mathcal{N}_v^r) : r \in \mathcal{R}\right\}\right)\right),
\end{equation}
where $\operatorname{Agg}_{r}(\cdot)$ is an edge-type-specific neighborhood aggregation function parameterized by $\boldsymbol{\theta}_r$, $\operatorname{Reduce}(\cdot)$ is a function to combine (e.g., sum) aggregation results from different edge types, and $\operatorname{Trans}_{\phi(v)}(\cdot)$ is a node-type-specific transformation parameterized by $\boldsymbol{\theta}_{\phi(v)}$.
Some HGNNs might contain additional schema-irrelevant parameters $\boldsymbol{\theta}_c$.
\end{definition}

\begin{definition}[Federated Graph Learning]
A federated graph learning (FGL) framework consists of a server and $K$ clients. The $k$-th client holds its own graph dataset $\mathcal{D}^k = \left(\mathcal{G}^k, \mathbf{X}^k[, \mathbf{Y}^k]\right)$, where $\mathcal{G}^k = \left(\mathcal{V}^k, \mathcal{E}^k\right)$ is the graph, $\mathbf{X}^k$ is the node feature matrix and $\mathbf{Y}^k$ is the optional label matrix.
The goal of FGL is to collaboratively train a shared graph model without disclosing clients' local data.
\end{definition}

\begin{definition}[Federated Heterogeneous Graph Learning]
Federated heterogeneous graph learning is a sub-category of FGL, where the graph $\mathcal{G}^k$ owned by each client is heterogeneous. The graph schema $\mathcal{T}_{\mathcal{G}^k}=(\mathcal{A}^k, \mathcal{R}^k)$ could be different or even kept private across clients.
\end{definition}



\begin{table}[tb]
    \centering
    \begin{tabular}{l|l}
        \toprule
        \multicolumn{1}{c}{\textbf{Notation}} & \multicolumn{1}{c}{\textbf{Definition}}\\
        \midrule
        $\mathbb{R}^{n}$ & $n$-dimensional Euclidean space\\
        $x$, $\mathbf{x}$, $\mathbf{X}$ & Scalar, vector, matrix\\
        $\mathcal{V}^k / \mathcal{E}^k$ & The set of nodes/edges at client $k$\\
        $\mathcal{G}^k$ & A graph $\mathcal{G}^k=\left(\mathcal{V}^k,\mathcal{E}^k\right)$ at client $k$\\
        $\mathcal{A} / \mathcal{R}$ & The set of node/edge types\\
        $\mathcal{N}_v^r$ & Type-$r$ edge connected neighbors of node $v$\\
        $\mathbf{h}_v$ & Hidden states (embedding) of node $v$\\
        $\boldsymbol{\theta}_s$ & Model parameters bound to node/edge type $s$\\
        $\boldsymbol{\theta}_c$ & Model parameters not bound to graph schema\\
        $\mathbf{v}$ & Basis weights (bases)\\
        $\boldsymbol{\beta}_s$ & Basis coefficients of node/edge type $s$ \\
        $|\cdot|$ & The cardinality of a set\\
        \bottomrule
    \end{tabular}
    \caption{Notations used in this paper.}
    \label{tab:notation}
\end{table}

\section{Methodology}

\begin{figure*}[t]
\centering
\includegraphics[width=0.9\textwidth]{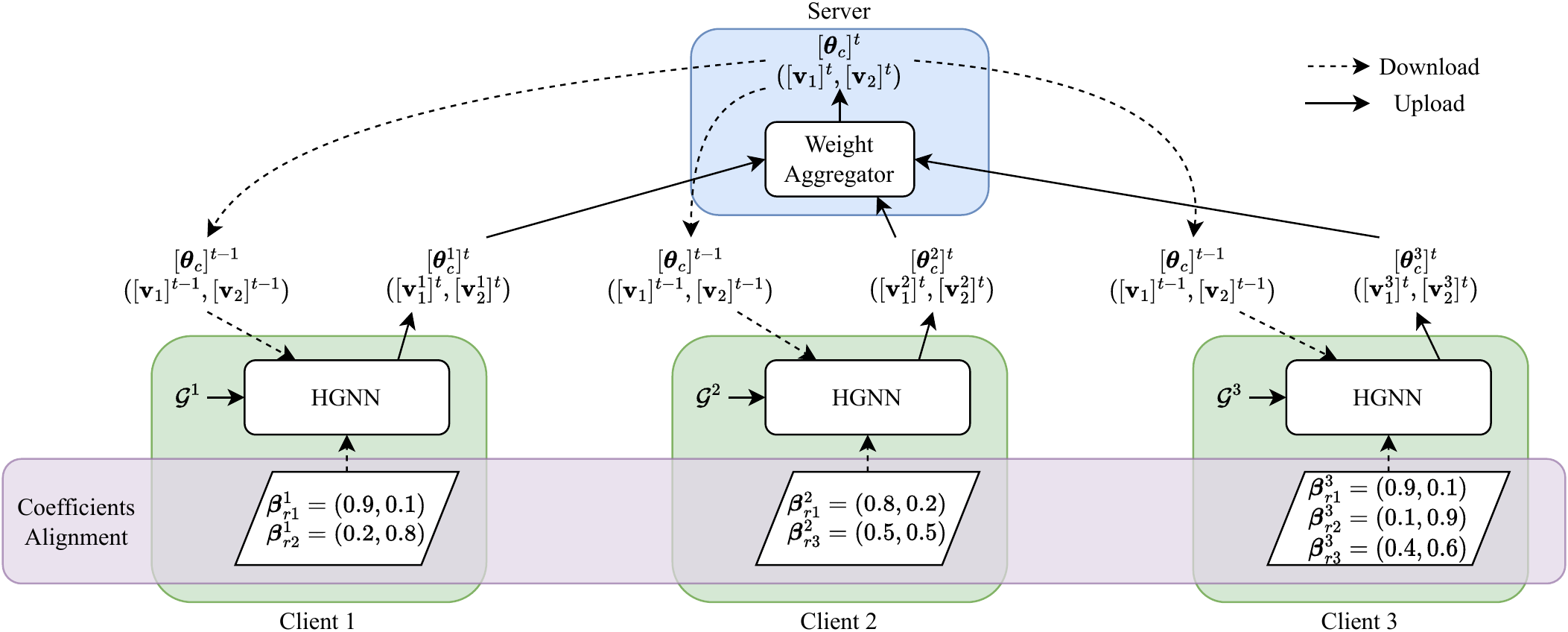} 
\caption{The overall architecture of FedHGN, with the number of clients $K=3$ and the number of bases $B=2$. Client 1 has edge types $r1$ and $r2$; client 2 has edge types $r1$ and $r3$; client 3 has edge types $r1$, $r2$, and $r3$. The uploads and downloads of $\boldsymbol{\beta}$ are not shown for brevity.}
\label{fig:FedHGN}
\end{figure*}

In this section, we propose FedHGN, a novel FGL framework for heterogeneous graphs and HGNNs.
FedHGN aims to achieve improved HGNN performance on each client through collaborative training under a federated framework without sharing graph data, node/edge attributes, and graph schema.
To attain this goal, FedHGN needs to separate graph schema from model weights before collecting and aggregating weights on the server side.
Therefore, we first propose a schema-weight decoupling (SWD) strategy to enable schema-agnostic knowledge sharing in Section~\ref{sec:swd}.
Then, to alleviate the adverse side effects introduced by the decoupling strategy, we design a coefficients alignment (CA) component in Section~\ref{sec:ca}.
Next, the overall FedHGN framework is depicted in Section~\ref{sec:sys}.
Finally, a privacy analysis of our proposed framework is conducted in Section~\ref{sec:privacy}.
A graphical illustration of FedHGN is provided in Figure~\ref{fig:FedHGN}.


\begin{figure}[t]
\centering
\includegraphics[width=0.8\columnwidth]{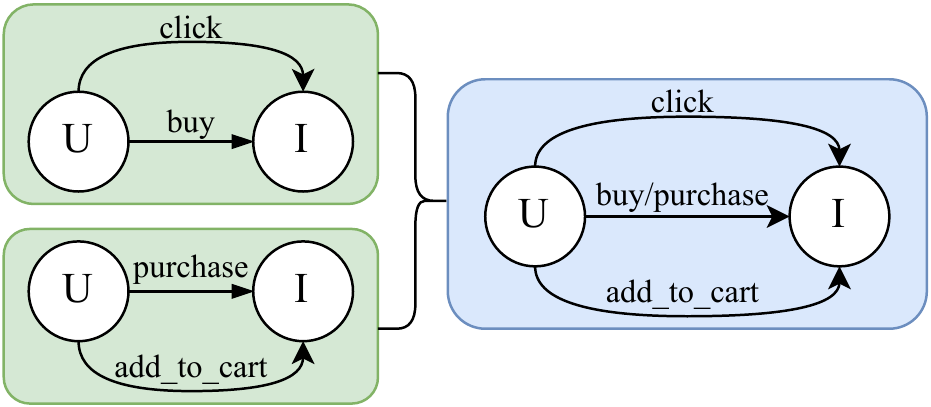}
\caption{An example of generating the global graph schema (blue) by matching local ones (green) from two clients. FedHGN does not require this privacy-sensitive operation.}
\label{fig:align}
\end{figure}

\subsection{Schema-Weight Decoupling} \label{sec:swd}

If naively combining HGNN and conventional FL methods (e.g., FedAvg), the server and the clients would need to first coordinate for a global graph schema.
Taking Figure~\ref{fig:align} as an example, the resulting global graph schema is generated by matching the two clients' local schemas, which inevitably breaches schema privacy and involves human expertise.

To avoid the privacy-sensitive schema matching process, we propose SWD to remove the correspondence between graph schemas and HGNN weights.
For schema-specific weights $\boldsymbol{\theta}_{s}^k \in \mathbb{R}^{d}$ of any node/edge type $s \in \mathcal{A}\cup\mathcal{R}$ from an HGNN at client $k$, one can apply basis decomposition by
\begin{equation}\label{eq:decouple}
    \boldsymbol{\theta}_{s}^k = \sum_{i=1}^{B} \beta_{s,i}^k\mathbf{v}_{i}^k,
\end{equation}
where $\beta_{s,i}^k \in \mathbb{R}$ are schema-specific coefficients, $\mathbf{v}_{i}^k \in \mathbb{R}^{d}$ are basis weights (or bases), and $B$ is a hyperparameter specifying the number of bases.
Here model weights are treated as flattened vectors for notational brevity. Without loss of generality, SWD can be easily extended to higher-dimensional weights.
The bases $\mathbf{v}_{i}^k$ are shared by all node/edge types, and not bound to the graph schema. They can be trained along with schema-irrelevant weights $\boldsymbol{\theta}_c^k$ under the conventional FL paradigm. 
On the other hand, a small number of coefficients $\boldsymbol{\beta}_{s}^k\in\mathbb{R}^{B}$ are schema-dependent. They are tuned locally at each client and do not participate in weight aggregation at the server.
By applying SWD and sharing only bases, FedHGN achieves schema-agnostic knowledge sharing, which avoids graph schema leakage and manual schema matching.

\subsection{Coefficients Alignment} \label{sec:ca}

Due to the randomness throughout training, the non-convex nature of the objective, and the non-IID heterogeneous graph data among clients, schema-specific coefficients $\boldsymbol{\beta}_{s}^k$ corresponding to the same node/edge type at different clients might diverge from each other towards distinct directions in the FL training process.
This discrepancy between schema coefficients among clients may lead to unstable training, prolonged convergence, and sub-optimal results.
Suppose the central server knows the matchings between schema coefficients among clients, this discrepancy issue can be easily resolved by aggregating and synchronizing them between the server and the clients.
However, this strong assumption violates schema privacy and requires schema matching, which is not allowed in federated heterogeneous graph learning.

To address the discrepancy issue without compromising the setting, we devise a heuristic approach to align the schema coefficients by enforcing a client-side loss constraint onto them.
Intuitively, even though schema coefficients of the same node/edge type at different clients may diverge in different directions,
there is a high chance of them being closer in the vector space than those of the different or dissimilar node/edge types.
During each communication round, in addition to the shareable bases $\mathbf{v}_{i}^k$ and schema-irrelevant weights $\boldsymbol{\theta}_c^k$, the FedHGN server also collects and distributes the schema-specific coefficients $\boldsymbol{\beta}_{s}^k$ from/to clients.
But instead of aggregating them on the server side, FedHGN introduces a novel regularization term on the client side to
penalize the local node/edge types' coefficients deviating from the most similar counterpart from other clients:
\begin{equation}\label{eq:align}
\begin{split}
\mathcal{L}_{align}^k = &\sum_{a\in\mathcal{A}^k}\min_{k^\prime \neq k,a^\prime} \left\lVert\boldsymbol{\beta}_{a}^{k} - \boldsymbol{\beta}_{a^\prime}^{k^\prime}\right\rVert_2^2 +\\
&\sum_{r\in\mathcal{R}^k}\min_{k^\prime \neq k,r^\prime} \left\lVert\boldsymbol{\beta}_{r}^{k} - \boldsymbol{\beta}_{r^\prime}^{k^\prime}\right\rVert_2^2,
\end{split}
\end{equation}
where $\boldsymbol{\beta}_{s}^k \in \mathbb{R}^B$ is the schema-specific coefficients of node/edge type $s \in \mathcal{A}^k\cup\mathcal{R}^k$ from client $k$'s HGNN model.
The key here is how to identify the most similar counterpart for $\boldsymbol{\beta}_{s}^{k}$.
We intuitively select the one with the smallest distance by taking the minimum, which has the highest chance of matching the node/edge type.
By minimizing this heuristic regularization term, FedHGN aligns schema coefficients across clients without disclosing actual graph schema.


\begin{algorithm}[t]
\caption{FedHGN framework.} \label{alg:FedHGN}
\begin{algorithmic}[1] 
\Require $K$ clients indexed by $k$; Client sampling fraction $C$; Number of bases $B$; Number of local epochs $E$; Learning rate $\eta$; Alignment regularization factor $\lambda$.
\Procedure{ServerExecution}{}
  \State Initialize $[\boldsymbol{\theta}_c]^{0}$, $[\mathbf{v}]^{0}$, and $\{[\boldsymbol{\beta}^{k}]^{0}: 1 \leq k \leq K\}$ (subscripts of $\mathbf{v}$ and $\boldsymbol{\beta}$ omitted for brevity)
  \For {round $t=1,2,\ldots$}
    \State $\mathcal{S}_t \leftarrow$ (random set of $\max(C\cdot K, 1)$ clients)
    \For {each client $k \in \mathcal{S}_t$ \textbf{in parallel}}
      \State $[\boldsymbol{\theta}_c^k]^{t}, [\mathbf{v}^k]^{t}, [\boldsymbol{\beta}^{k}]^{t} \leftarrow \operatorname{ClientUpdate}($
      \Statex[4] $k, [\boldsymbol{\theta}_c]^{t-1}, [\mathbf{v}]^{t-1}, \{[\boldsymbol{\beta}^{k^\prime}]^{t-1}:k^\prime \neq k\})$
    \EndFor
    \State $([\boldsymbol{\theta}_c]^{t},[\mathbf{v}]^t) \leftarrow \sum_{k \in \mathcal{S}_t}\frac{N_k}{N_{\mathcal{S}_t}} ([\boldsymbol{\theta}_c^k]^{t},[\mathbf{v}^k]^t)$
    \State $[\boldsymbol{\beta}^{k}]^{t} \leftarrow [\boldsymbol{\beta}^{k}]^{t-1}, \forall k \notin \mathcal{S}_t$
  \EndFor
\EndProcedure
\Procedure{ClientUpdate}{$k$, $\boldsymbol{\theta}_c$, $\mathbf{v}$, $\{\boldsymbol{\beta}^{k^\prime}:k^\prime \neq k\}$}
  \State $\boldsymbol{\theta}_{s} \leftarrow \sum_{i=1}^{B} \beta_{s,i}^k\mathbf{v}_{i}, \forall s\in\mathcal{A}^k\cup\mathcal{R}^k$
  \For {epoch $i=1,2,\ldots,E$}
    \For {minibatch $m$ sampled from $\mathcal{D}^k$}
      \State $\mathcal{L}_{task}^k \leftarrow$ (task-specific loss given $m$)
      \State $\mathcal{L}_{align}^k \leftarrow$ Equation~(\ref{eq:align})
      \State $\boldsymbol{\theta}_c, \mathbf{v}, \boldsymbol{\beta}^k \leftarrow (\boldsymbol{\theta}_c, \mathbf{v}, \boldsymbol{\beta}^k) - \eta\nabla\left(\mathcal{L}_{task}^k + \lambda \mathcal{L}_{align}^k\right)$
    \EndFor
  \EndFor
  \State Return $\boldsymbol{\theta}_c$, $\mathbf{v}$, and $\boldsymbol{\beta}^k$ to server
\EndProcedure
\end{algorithmic}
\end{algorithm}

\subsection{Framework Overview} \label{sec:sys}

Our FedHGN framework largely follows FedAvg, which iteratively conducts local training at multiple clients and weight aggregation at a central server.
In addition to the typical operations of FedAvg, FedHGN employs SWD and CA to preserve schema privacy without performance degradation.
Before starting the federated training process, the server and the clients would negotiate for the hyperparameters of the HGNN model used, such as the number of bases ($B$) in SWD.
The primary training process comprises many communication rounds between the server and the clients.
The detailed server-side and client-side operations of a single round in FedHGN are described below.
Figure~\ref{fig:FedHGN} illustrates this training process.
Complete pseudocode is given in Algorithm~\ref{alg:FedHGN}.

\paragraph{Server Operations.} At round $t$, the server first samples a fraction of clients $\mathcal{S}_t$ for this round of training.
The server then sends the latest model parameters to each client $k \in \mathcal{S}_t$, including the aggregated schema-agnostic weights, $[\boldsymbol{\theta}_c]^{t-1}$ and $[\mathbf{v}_{i}]^{t-1}$, and collected schema coefficients $[\boldsymbol{\beta}_{s}^{k^\prime}]^{t-1}$ from the previous round.
After client-side local updates, the server receives the updated $[\boldsymbol{\theta}_c^k]^{t}$, $[\mathbf{v}_{i}^k]^{t}$, and $[\boldsymbol{\beta}_{s}^{k}]^{t}$ from each $k \in \mathcal{S}_t$.
Finally, the server aggregates the schema-agnostic weights by $[\boldsymbol{\theta}_c]^{t} = \sum_{k \in \mathcal{S}_t}\frac{N_k}{N_{\mathcal{S}_t}} [\boldsymbol{\theta}_c^k]^{t}$ and $[\mathbf{v}_{i}]^t = \sum_{k \in \mathcal{S}_t}\frac{N_k}{N_{\mathcal{S}_t}} [\mathbf{v}_{i}^k]^t$,
where $N_k$ is the number of training samples at client $k$, and $N_{\mathcal{S}_t} = \sum_{k \in \mathcal{S}_t}N_k$ is the total number of training samples of this round.
The server also sets $[\boldsymbol{\beta}_{s}^{k}]^{t} = [\boldsymbol{\beta}_{s}^{k}]^{t-1}$ for $k \notin \mathcal{S}_t$.

\paragraph{Client Operations.} At round $t$, each selected client $k$ receives the latest model parameters from the server, including schema-agnostic weights, $[\boldsymbol{\theta}_c]^{t-1}$ and $[\mathbf{v}_{i}]^{t-1}$, and other clients' schema coefficients $[\boldsymbol{\beta}_{s}^{k^\prime}]^{t-1}$.
The schema-specific weights are recreated via Equation~(\ref{eq:decouple}).
Then, the client would conduct local training using local heterogeneous graph data $\mathcal{D}^k$ for several epochs to tune model parameters. The client-side objective function is a sum of the task-specific loss and the CA regularization term: $\mathcal{L}^k = \mathcal{L}_{task}^k + \lambda \mathcal{L}_{align}^k$, where $\mathcal{L}_{task}^k$ could be cross entropy for node classification tasks or some ranking loss for link prediction tasks.
Finally, the client uploads the updated schema-agnostic HGNN weights, $[\boldsymbol{\theta}_c^k]^{t}$ and $[\mathbf{v}_{i}^k]^{t}$, and schema coefficients $[\boldsymbol{\beta}_{s}^{k}]^{t}$ back to the server.


\subsection{Privacy Analysis} \label{sec:privacy}
In this section, we analyze the communications between the server and the clients and discuss whether FedHGN can preserve the data privacy of local graph schemas.

\paragraph{Server-side Analysis.}
FedHGN server is restricted to collecting limited information that cannot be used to infer graph schemas of any clients.
The central server receives only HGNN model weights $\mathbf{v}_{i}^k$, $\boldsymbol{\theta}_c^k$, and $\boldsymbol{\beta}_{s}^{k}$ from each client.
$\mathbf{v}_{i}^k$ and $\boldsymbol{\theta}_c^k$ are schema-agnostic model weights, revealing no schema information.
Although $\boldsymbol{\beta}_{s}^{k}$ are schema-specific coefficients, the server can only deduce the number of node/edge types in each client. It still cannot learn other meaningful schema information, like the matching among them or the physical name of their corresponding node/edge types.


\paragraph{Client-side Analysis.}
Data privacy at each client is well protected under the FedHGN framework.
Each client $k$ receives $\mathbf{v}_{i}$, $\boldsymbol{\theta}_c$, and $\boldsymbol{\beta}_{s}^{k^\prime}$ (for $k^\prime \neq k$) from the server.
Similarly, clients cannot infer other clients' graph schemas from $\mathbf{v}_{i}$ and $\boldsymbol{\theta}_c$.
Since the schema-specific coefficients $\boldsymbol{\beta}_{s}^{k^\prime}$ are sent to client $k$ as an unordered set, client $k$ cannot differentiate the source client of each received $\boldsymbol{\beta}_{s}^{k^\prime}$.
Therefore, neither the HGNN model nor the graph schema of any other client can be derived based on what client $k$ receives from the server.


\begin{table}[t]
\centering
\begin{tabular}{@{}clrrr@{}}
\toprule
\multicolumn{2}{c}{$K$}                               & \multicolumn{1}{c}{3} & \multicolumn{1}{c}{5} & \multicolumn{1}{c}{10} \\ \midrule
\multicolumn{1}{c|}{\multirow{4}{*}{RE}}  & \# ntypes & 27.0                  & 27.0                  & 27.0                   \\
\multicolumn{1}{c|}{}                     & \# etypes & 120.7                 & 122.0                 & 120.0                  \\
\multicolumn{1}{c|}{}                     & \# nodes  & 80,592.0              & 68,697.6              & 50,238.5               \\
\multicolumn{1}{c|}{}                     & \# edges  & 358,873.3             & 249,948.4             & 142,849.6              \\ \midrule
\multicolumn{1}{c|}{\multirow{4}{*}{RET}} & \# ntypes & 20.3                  & 18.0                  & 12.3                   \\
\multicolumn{1}{c|}{}                     & \# etypes & 66.0                  & 52.4                  & 26.8                   \\
\multicolumn{1}{c|}{}                     & \# nodes  & 77,027.7              & 73,202.4              & 43,735.6               \\
\multicolumn{1}{c|}{}                     & \# edges  & 349,813.3             & 248,125.6             & 138,807.0              \\ \bottomrule
\end{tabular}
\caption{Averaged statistics of the BGS dataset split into $K$ clients.}
\label{tab:dataset}
\end{table}

\begin{table*}[t]
\centering
\begin{tabular}{@{}ccccccccc@{}}
\toprule
\multicolumn{2}{c}{Dataset}                         & \multicolumn{3}{c}{AIFB}                                                    & \multicolumn{2}{c}{MUTAG}                         & \multicolumn{2}{c}{BGS}                           \\ \midrule
\multicolumn{2}{c}{Central}                         & \multicolumn{3}{c}{87.78$\pm$2.22}                                          & \multicolumn{2}{c}{68.24$\pm$3.79}                & \multicolumn{2}{c}{81.38$\pm$1.69}                \\ \midrule
\multicolumn{2}{c}{$K$}                             & 3                       & 5                       & 10                      & 5                       & 10                      & 5                       & 10                      \\ \midrule
\multicolumn{1}{c|}{\multirow{4}{*}{RE}}  & Local   & {\ul 74.77$\pm$5.22}    & {\ul 71.30$\pm$4.30}    & {\ul 55.08$\pm$1.09}    & 65.06$\pm$1.28          & 64.88$\pm$1.12          & 67.04$\pm$1.79          & 64.54$\pm$1.65          \\
\multicolumn{1}{c|}{}                     & FedAvg  & 74.02$\pm$3.09          & 65.18$\pm$3.59          & 54.95$\pm$1.42          & 65.06$\pm$0.34          & 65.47$\pm$0.74          & {\ul 67.36$\pm$2.52}    & {\ul 65.04$\pm$1.26}    \\
\multicolumn{1}{c|}{}                     & FedProx & 72.34$\pm$3.37          & 65.29$\pm$5.43          & 53.38$\pm$2.79          & {\ul 65.59$\pm$0.69}    & \textbf{65.62$\pm$0.76} & 66.11$\pm$3.99          & 63.97$\pm$1.66          \\
\multicolumn{1}{c|}{}                     & FedHGN  & \textbf{81.87$\pm$2.33} & \textbf{72.94$\pm$3.27} & \textbf{59.32$\pm$1.42} & \textbf{67.87$\pm$0.83} & {\ul 65.59$\pm$0.86}    & \textbf{67.52$\pm$2.58} & \textbf{65.73$\pm$1.47} \\ \midrule
\multicolumn{1}{c|}{\multirow{4}{*}{RET}} & Local   & 76.11$\pm$3.58          & 65.89$\pm$2.29          & 59.71$\pm$3.06          & 64.78$\pm$0.89          & 64.18$\pm$3.13          & 67.73$\pm$1.19          & {\ul 67.11$\pm$0.52}    \\
\multicolumn{1}{c|}{}                     & FedAvg  & 80.00$\pm$1.72          & {\ul 66.56$\pm$3.47}    & 59.59$\pm$3.03          & {\ul 66.04$\pm$0.11}    & \textbf{66.28$\pm$0.15} & 69.10$\pm$2.11          & 65.04$\pm$0.96          \\
\multicolumn{1}{c|}{}                     & FedProx & {\ul 80.93$\pm$2.78}    & 66.00$\pm$2.95          & {\ul 60.99$\pm$1.95}    & 65.92$\pm$0.29          & 66.13$\pm$0.06          & \textbf{70.62$\pm$1.42} & 66.07$\pm$0.80          \\
\multicolumn{1}{c|}{}                     & FedHGN  & \textbf{86.55$\pm$2.53} & \textbf{70.67$\pm$2.42} & \textbf{66.76$\pm$3.22} & \textbf{66.21$\pm$0.40} & {\ul 66.21$\pm$0.12}    & {\ul 69.79$\pm$2.07}    & \textbf{69.03$\pm$1.44} \\ \bottomrule
\end{tabular}
\caption{Experimental results (weighted average of accuracy \%) of node classification.}
\label{tab:node_classification}
\end{table*}

\section{Experiments}

In this section, we demonstrate the effectiveness of FedHGN on federated learning of HGNNs
by conducting experiments for \emph{node classification} on a series of heterogeneous graph datasets.
The experiments are designed to answer the following research questions.
\textbf{RQ1}: Can FedHGN achieve a better HGNN performance than purely local training?
\textbf{RQ2}: Can FedHGN outperform conventional FL methods in training HGNNs?
\textbf{RQ3}: What are the effects of the proposed schema-weight decoupling and coefficients alignment?
\textbf{RQ4}: How is FedHGN affected by the hyperparameters?


\subsection{Experimental Setup}

\paragraph{Datasets.} We select widely-adopted heterogeneous graph datasets for our node classification experiments: AIFB, MUTAG, and BGS preprocessed by Deep Graph Library (DGL)~\cite{DBLP:journals/corr/abs-1909-01315}.
These datasets are highly heterogeneous with more than 50 node/edge types.
To simulate the federated setting, we randomly split each dataset into $K=3$, $5$, and $10$ clients.
We propose two random splitting strategies for heterogeneous graphs:
(1) \textbf{Random Edges (RE)} which randomly allocates edges to $K$ clients and
(2) \textbf{Random Edge Types (RET)} which randomly allocates edge types to $K$ clients.
RE simulates a scenario where each client owns a part of the complete graph, within which there are overlapping nodes/edges among clients.
RET simulates a more complex scenario where clients construct the local heterogeneous graphs differently, resulting in distinct graph schemas.
Averaged statistics of the BGS dataset split into $K$ clients are summarized in Table~\ref{tab:dataset}.
More details about the datasets and splitting strategies are provided in the appendix.

\paragraph{Baselines.} To assess FedHGN and demonstrate its superiority, we choose the following settings/algorithms as baselines: (1) centralized training, (2) local training, (3) FedAvg, and (4) FedProx.
We do not include existing FGL methods as baselines because they cannot be easily adapted to HGNNs.
FedAlign is not adopted because its source code is not available and some design details are not disclosed in its paper.


\paragraph{Implementation.}
We employ an RGCN-like architecture as the HGNN model. 
Other details are in the appendix.

\subsection{Experimental Results}

We report results averaged from 5 runs using different random seeds.
The accuracy score of each run is computed via a weighted average of the $K$ clients' local testing scores based on the numbers of their testing samples.

\subsubsection{Main Results (RQ1 \& RQ2)}

To evaluate FedHGN against local training and conventional FL algorithms, we conduct comprehensive experiments on three datasets with varying client numbers.
Note that schema sharing is required and allowed in FedAvg and FedProx, but not in FedHGN.
Averaged accuracy scores with standard deviations are listed in Table~\ref{tab:node_classification}.
Due to the space limit, we leave the $K=3$ settings of MUTAG and BGS in the appendix.

As shown in the table, FedHGN consistently outperforms local training and conventional FL algorithms in most settings.
As expected, HGNN performance deteriorates in local training due to insufficient data, especially on AIFB and BGS.
But conventional FL algorithms (FedAvg and FedProx) do not have a noticeable advantage over local training in our experiments. On AIFB (RE), they are even outperformed by the local setting.
Although FedProx is designed to alleviate the non-IID data issue, it cannot correctly handle heterogeneous graph data based on our results.
Compared to FedAvg and FedProx, FedHGN makes no compromise in schema-level privacy and yet achieves superior results, showing that FedHGN is a better option for federated HGNNs.
FedHGN's advantage in preserving HGNN performance might come from the decoupled schema coefficients.
They may serve as personalized model parameters to help each client combat the severe non-IID problem of heterogeneous graphs.




\begin{table}[t]
\centering
\begin{tabular}{@{}cccccc@{}}
\toprule
\multirow{2}{*}{SWD} & \multirow{2}{*}{CA} & \multicolumn{2}{c}{MUTAG (RE)}                           & \multicolumn{2}{c}{BGS (RE)}                                \\ \cmidrule(l){3-6} 
                     &                     & \multicolumn{1}{c}{$K=5$} & $K=10$                  & \multicolumn{1}{c}{$K=5$} & \multicolumn{1}{c}{$K=10$} \\ \midrule
B                    & \xmark              & 63.18            & 60.59          & 66.39            & 59.36             \\
C                    & \xmark              & {\ul 65.88}      & 61.26          & \textbf{67.78}   & 64.11             \\
B+C                  & \xmark              & 65.29            & 62.83          & 66.53            & 63.76             \\
\xmark               & \xmark              & 65.06            & {\ul 65.47}    & 67.36            & {\ul 65.04}       \\ \midrule
B                    & \cmark              & \textbf{67.87}   & \textbf{65.59} & {\ul 67.52}      & \textbf{65.73}    \\ \bottomrule
\end{tabular}
\caption{Ablation study on the MUTAG and BGS datasets (RE splitting strategy). B: aggregating basis weights. C: aggregating schema-specific coefficients. The last row is our proposed FedHGN.}
\label{tab:ablation}
\end{table}

\subsubsection{Ablation Study (RQ3)}

To analyze the effects of the proposed designs, we compare variants of FedHGN on MUTAG (RE) and BGS (RE) in Table~\ref{tab:ablation} by
applying different \emph{schema-weight decoupling} (SWD) strategies
and whether employing \emph{coefficients alignment} (CA) or not.
The table shows that CA brings about a 7\% improvement compared to the variant without it.
We also notice a longer convergence time when removing CA.
Therefore, the alignment component is necessary for FedHGN to stabilize training and maintain performance.
For SWD, we observe that additionally aggregating schema coefficients at the server side (breaking schema privacy) leads to some performance gain, which can be well compensated by CA.
Another interesting observation is that the FedHGN variant that only aggregates schema coefficients achieves decent results. We suspect it may resemble the local setting because the shared coefficients have a small parameter space.
We also present the results produced by removing both SWD and CA, which reduces FedHGN to FedAvg. Although outperforming the variant that uses only SWD, FedAvg violates schema privacy and is still inferior to the full version of FedHGN.
All these results confirm the effectiveness of SWD and CA.

\subsubsection{Hyperparameter Analysis (RQ4)}

\begin{figure}[t]
\centering
    \begin{subfigure}[b]{0.49\columnwidth}
        \centering
        \includegraphics[width=\linewidth]{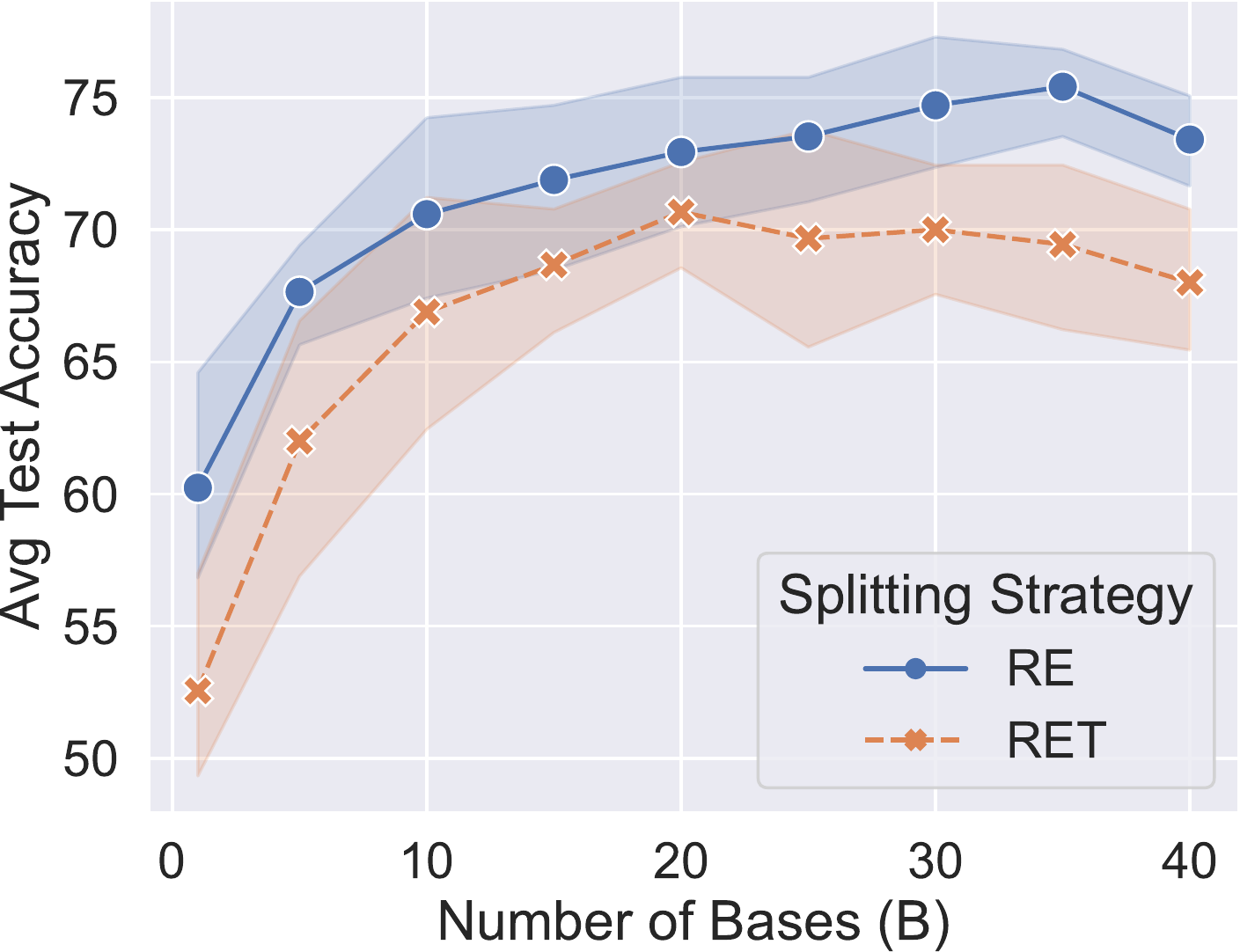}
        \caption{The number of bases ($B$).}
    \end{subfigure}
    \hfill
    \begin{subfigure}[b]{0.49\columnwidth}
        \centering
        \includegraphics[width=\linewidth]{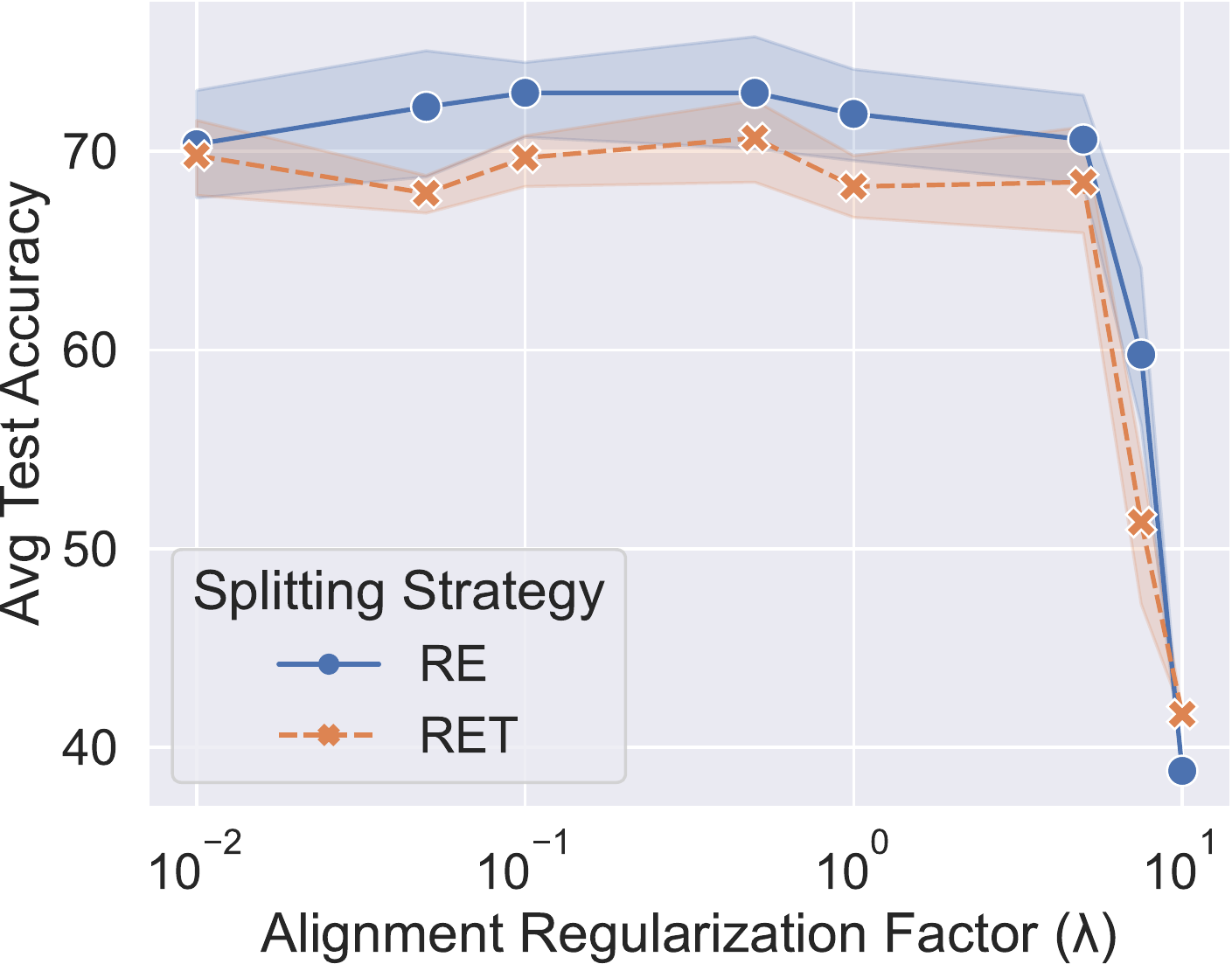}
        \caption{The alignment factor ($\lambda$).}
    \end{subfigure}
\caption{Hyperparameter analysis of varying $B$ and $\lambda$.}
\label{fig:hyperparam}
\end{figure}

To investigate how FedHGN is affected by the newly introduced hyperparameters, we conduct pilot tests on AIFB with different choices for the number of bases $B$ and the alignment regularization factor $\lambda$. Except for the concerning hyperparameter, all other setups are kept unchanged.
From Figure~\ref{fig:hyperparam}, we find that FedHGN is sensitive to the choice of $B$. The optimal performance is reached at around $B=35$ and $B=20$ for AIFB (RE) and AIFB (RET), respectively.
Intuitively, $B$ should be related to the number of edge types involved in the FL system.
Since other settings (MUTAG and BGS) contain a similar or smaller number of edge types, we roughly choose $B=20$ as a sugar point to report the main results.

For the alignment regularization factor $\lambda$, FedHGN performance does not change much when $\lambda$ is reasonable, i.e., for $\lambda \leq 5$.
But when $\lambda$ gets larger than 5, the testing accuracy drops dramatically on both AIFB (RE) and AIFB (RET).
When $\lambda$ is too large, the optimization objective is dominated by the CA regularization term $\mathcal{L}_{align}$. HGNNs cannot learn much knowledge from the task-specific loss $\mathcal{L}_{task}$ in this case.
Hence, we just set $\lambda=0.5$ for our main experiments.


\section{Discussion}

This work aims to avoid schema matching and ensure schema privacy during federated HGNN learning.
By carefully examining the HGNN architecture, we notice the crux of the matter is the correspondence between model weights and node/edge types.
Our FedHGN is designed to decouple this binding and achieve schema-agnostic weight sharing.
This seems unnecessary as one could just apply secure multiparty computation (MPC) techniques like private set operations~\cite{DBLP:conf/eurocrypt/FreedmanNP04,DBLP:conf/crypto/KissnerS05} to match the node/edge types across clients.
However, node/edge types' naming conventions may differ across clients.
Human labor is usually inevitable to coordinate a set of naming rules, which generally breaks schema privacy.
Furthermore, without SWD, the server can derive client schema from received gradient/weight changes between communication rounds.
Although gradients can be protected by
differential privacy~\cite{DBLP:conf/ccs/AbadiCGMMT016}, homomorphic encryption~\cite{DBLP:journals/tifs/PhongAHWM18}, secure aggregation~\cite{DBLP:conf/ccs/BonawitzIKMMPRS17}, or MPC~\cite{DBLP:conf/sp/MohasselZ17},
these approaches could affect model performance, increase computational overhead, or impede communication reduction methods~\cite{DBLP:conf/globecom/ErgunSG22}.
Therefore, we reckon FedHGN as a practical design to reduce schema privacy risks.




\section{Conclusion}


This paper proposes FedHGN, a novel federated heterogeneous graph learning framework for HGNNs with schema privacy protection.
By decoupling graph schemas and model weights, FedHGN removes the correspondence between node/edge types and HGNN weights to achieve schema-agnostic knowledge sharing.
With an alignment regularization term, the discrepancy among same-type schema coefficients across clients is alleviated to stabilize training and boost model performance.
Experimental results show that FedHGN can bring consistent performance improvement to HGNNs while preserving both data and schema privacy, outperforming local training and conventional FL algorithms.


\section*{Acknowledgments}

The work described here was partially supported by grants from the National Key Research and Development Program of China (No. 2018AAA0100204) and from the Research Grants Council of the Hong Kong Special Administrative Region, China (RGC GRF No. 2151185, CUHK 14222922).

\bibliographystyle{named}
\bibliography{ijcai23}

\newpage
\appendix

\section{Datasets}

We select widely-adopted heterogeneous graph datasets for our node classification experiments.
Specifically, we adopt three datasets in Resource Description Framework (RDF) format: \textbf{AIFB}, \textbf{MUTAG}, and \textbf{BGS} preprocessed by Deep Graph Library (DGL)~\cite{DBLP:journals/corr/abs-1909-01315}.

To simulate a federated training setting, we propose two random splitting strategies as follows.
Averaged statistics of the datasets split into $K$ clients are summarized in Table~\ref{tab:dataset_full}.
\begin{itemize}
    \item \textbf{Random edges (RE)}. The edges of the given graph are randomly and evenly divided into $K+2$ groups. Each client exclusively possesses one group of edges. All clients share another group of edges. The edges in the remaining group are distributed to randomly selected $p$ ($1<p<K$) clients.
    \item \textbf{Random edge types (RET)}. The edge types of the given graph are randomly and evenly divided into $K+2$ groups. Each client exclusively possesses one group of edge types. All clients share another group of edge types. The edge types in the remaining group are distributed to randomly selected $p$ ($1<p<K$) clients.
\end{itemize}

\begin{table*}[t]
\centering
\begin{tabular}{@{}clrrrrrrrrr@{}}
\toprule
\multicolumn{2}{c}{Dataset}                           & \multicolumn{3}{c}{AIFB}                                               & \multicolumn{3}{c}{MUTAG}                                              & \multicolumn{3}{c}{BGS}                                                \\ \midrule
\multicolumn{2}{c}{$K$}                               & \multicolumn{1}{c}{3} & \multicolumn{1}{c}{5} & \multicolumn{1}{c}{10} & \multicolumn{1}{c}{3} & \multicolumn{1}{c}{5} & \multicolumn{1}{c}{10} & \multicolumn{1}{c}{3} & \multicolumn{1}{c}{5} & \multicolumn{1}{c}{10} \\ \midrule
\multicolumn{1}{c|}{\multirow{7}{*}{RE}}  & \# ntypes & 7.0                   & 7.0                   & 7.0                    & 5.0                   & 5.0                   & 5.0                    & 27.0                  & 27.0                  & 27.0                   \\
\multicolumn{1}{c|}{}                     & \# etypes & 104.0                 & 103.2                 & 101.6                  & 50.0                  & 50.0                  & 49.0                   & 120.7                 & 122.0                 & 120.0                  \\
\multicolumn{1}{c|}{}                     & \# nodes  & 4,818.7               & 3,916.8               & 2,941.3                & 22,969.0              & 19,961.2              & 14,661.6               & 80,592.0              & 68,697.6              & 50,238.5               \\
\multicolumn{1}{c|}{}                     & \# edges  & 26,032.0              & 18,140.8              & 10,390.6               & 78,986.7              & 54,976.8              & 31,489.8               & 358,873.3             & 249,948.4             & 142,849.6              \\
\multicolumn{1}{c|}{}                     & \# train  & 110.7                 & 105.8                 & 94.1                   & 218.0                 & 217.8                 & 217.1                  & 94.0                  & 93.8                  & 92.6                   \\
\multicolumn{1}{c|}{}                     & \# valid  & 27.7                  & 27.8                  & 25.2                   & 54.0                  & 54.0                  & 54.0                   & 23.0                  & 23.0                  & 22.6                   \\
\multicolumn{1}{c|}{}                     & \# test   & 35.7                  & 34.0                  & 29.3                   & 68.0                  & 68.0                  & 68.0                   & 29.0                  & 28.8                  & 28.2                   \\ \midrule
\multicolumn{1}{c|}{\multirow{7}{*}{RET}} & \# ntypes & 7.0                   & 6.4                   & 6.0                    & 4.3                   & 4.2                   & 3.7                    & 20.3                  & 18.0                  & 12.3                   \\
\multicolumn{1}{c|}{}                     & \# etypes & 57.3                  & 43.2                  & 28.8                   & 26.7                  & 23.6                  & 13.0                   & 66.0                  & 52.4                  & 26.8                   \\
\multicolumn{1}{c|}{}                     & \# nodes  & 4,600.0               & 3,936.2               & 2,632.0                & 20,185.7              & 24,098.4              & 19,670.1               & 77,027.7              & 73,202.4              & 43,735.6               \\
\multicolumn{1}{c|}{}                     & \# edges  & 26,110.0              & 22,079.2              & 17,115.6               & 77,898.7              & 76,971.2              & 49,366.6               & 349,813.3             & 248,125.6             & 138,807.0              \\
\multicolumn{1}{c|}{}                     & \# train  & 112.0                 & 112.0                 & 107.9                  & 218.0                 & 212.4                 & 207.1                  & 94.0                  & 94.0                  & 94.0                   \\
\multicolumn{1}{c|}{}                     & \# valid  & 28.0                  & 28.0                  & 26.9                   & 54.0                  & 53.0                  & 52.4                   & 23.0                  & 23.0                  & 23.0                   \\
\multicolumn{1}{c|}{}                     & \# test   & 36.0                  & 36.0                  & 34.2                   & 68.0                  & 67.0                  & 66.2                   & 29.0                  & 29.0                  & 29.0                   \\ \bottomrule
\end{tabular}
\caption{Averaged statistics of the datasets after being split into $K$ clients.}
\label{tab:dataset_full}
\end{table*}

\section{Baselines}

To assess FedHGN and fully demonstrate its superiority, we choose the following three types of settings/algorithms as baselines: (1) centralized training, (2) local training, and (3) two conventional FL algorithms: FedAvg~\cite{DBLP:conf/aistats/McMahanMRHA17} and FedProx~\cite{DBLP:conf/mlsys/LiSZSTS20}.
\begin{itemize}
    \item \textbf{Central}. The HGNN is trained using the complete graph without data splitting.
    \item \textbf{Local}. The HGNN is trained separately for each client using only local heterogeneous graph data.
    \item \textbf{FedAvg}. The HGNN is trained using FedAvg without schema-weight decoupling and coefficients alignment. This FL setting assumes graph schemas are shared and matched, breaching schema privacy.
    \item \textbf{FedProx}. This one is similar to the FedAvg setting above. The loss function contains a proximal term to reduce the influence of system and data heterogeneity. This FL setting assumes graph schemas are shared and matched, breaching schema privacy.
\end{itemize}

\section{Implementation Details}

We employ an RGCN-like architecture as the HGNN model, which ignores node types and decouples edge-type-specific weights.
Throughout the experiments, unless otherwise specified, we fix
the client sampling fraction $C=1.0$,
the embedding dimension $d=64$,
the number of HGNN layers $L=2$,
the number of bases $B=20$,
the alignment regularization factor $\lambda=0.5$,
the number of local epochs $E=3$,
and the early stopping patience as 10 rounds.
We perform full-batch training using a stochastic gradient descent (SGD) optimizer with a learning rate $\eta=0.1$.
For baselines, we keep applicable hyperparameters to be the same as above for a fair comparison.
We implement FedHGN and all baselines using the PyTorch and DGL packages, and conduct all experiments on one NVIDIA RTX 3090 GPU.

\section{More Experimental Results}

Due to the space limit of the main body, we provide the complete experimental results here.
Results on the AIFB dataset are provided in Table~\ref{tab:node_classification_AIFB}.
Results on the MUTAG dataset are provided in Table~\ref{tab:node_classification_MUTAG}.
Results on the BGS dataset are provided in Table~\ref{tab:node_classification_BGS}.
Ablation study results with standard deviations are provided in Table~\ref{tab:ablation_full}.

\begin{table}[t]
\centering
\begin{tabular}{@{}ccccc@{}}
\toprule
\multicolumn{2}{c}{Dataset}                         & \multicolumn{3}{c}{AIFB}                                                    \\ \midrule
\multicolumn{2}{c}{Central}                         & \multicolumn{3}{c}{87.78$\pm$2.22}                                          \\ \midrule
\multicolumn{2}{c}{$K$}                             & 3                       & 5                       & 10                      \\ \midrule
\multicolumn{1}{c|}{\multirow{4}{*}{RE}}  & Local   & {\ul 74.77$\pm$5.22}    & {\ul 71.30$\pm$4.30}    & {\ul 55.08$\pm$1.09}    \\
\multicolumn{1}{c|}{}                     & FedAvg  & 74.02$\pm$3.09          & 65.18$\pm$3.59          & 54.95$\pm$1.42          \\
\multicolumn{1}{c|}{}                     & FedProx & 72.34$\pm$3.37          & 65.29$\pm$5.43          & 53.38$\pm$2.79          \\
\multicolumn{1}{c|}{}                     & FedHGN  & \textbf{81.87$\pm$2.33} & \textbf{72.94$\pm$3.27} & \textbf{59.32$\pm$1.42} \\ \midrule
\multicolumn{1}{c|}{\multirow{4}{*}{RET}} & Local   & 76.11$\pm$3.58          & 65.89$\pm$2.29          & 59.71$\pm$3.06          \\
\multicolumn{1}{c|}{}                     & FedAvg  & 80.00$\pm$1.72          & {\ul 66.56$\pm$3.47}    & 59.59$\pm$3.03          \\
\multicolumn{1}{c|}{}                     & FedProx & {\ul 80.93$\pm$2.78}    & 66.00$\pm$2.95          & {\ul 60.99$\pm$1.95}    \\
\multicolumn{1}{c|}{}                     & FedHGN  & \textbf{86.55$\pm$2.53} & \textbf{70.67$\pm$2.42} & \textbf{66.76$\pm$3.22} \\ \bottomrule
\end{tabular}
\caption{Experimental results (weighted average of accuracy \%) of node classification on the AIFB dataset.}
\label{tab:node_classification_AIFB}
\end{table}

\begin{table}[t]
\centering
\begin{tabular}{@{}cclcc@{}}
\toprule
\multicolumn{2}{c}{Dataset}                         & \multicolumn{3}{c}{MUTAG}                                                   \\ \midrule
\multicolumn{2}{c}{Central}                         & \multicolumn{3}{c}{68.24$\pm$3.79}                                          \\ \midrule
\multicolumn{2}{c}{$K$}                             & \multicolumn{1}{c}{3}   & 5                       & 10                      \\ \midrule
\multicolumn{1}{c|}{\multirow{4}{*}{RE}}  & Local   & 63.53$\pm$1.77          & 65.06$\pm$1.28          & 64.88$\pm$1.12          \\
\multicolumn{1}{c|}{}                     & FedAvg  & 64.42$\pm$1.18          & 65.06$\pm$0.34          & 65.47$\pm$0.74          \\
\multicolumn{1}{c|}{}                     & FedProx & 66.08$\pm$3.46          & {\ul 65.59$\pm$0.69}    & \textbf{65.62$\pm$0.76} \\
\multicolumn{1}{c|}{}                     & FedHGN  & \textbf{67.06$\pm$1.26} & \textbf{67.87$\pm$0.83} & {\ul 65.59$\pm$0.86}    \\ \midrule
\multicolumn{1}{c|}{\multirow{4}{*}{RET}} & Local   & 63.24$\pm$0.62          & 64.78$\pm$0.89          & 64.18$\pm$3.13          \\
\multicolumn{1}{c|}{}                     & FedAvg  & 64.61$\pm$1.33          & {\ul 66.04$\pm$0.11}    & \textbf{66.28$\pm$0.15} \\
\multicolumn{1}{c|}{}                     & FedProx & 65.30$\pm$2.45          & 65.92$\pm$0.29          & 66.13$\pm$0.06          \\
\multicolumn{1}{c|}{}                     & FedHGN  & \textbf{65.98$\pm$2.86} & \textbf{66.21$\pm$0.40} & {\ul 66.21$\pm$0.12}    \\ \bottomrule
\end{tabular}
\caption{Experimental results (weighted average of accuracy \%) of node classification on the MUTAG dataset.}
\label{tab:node_classification_MUTAG}
\end{table}

\begin{table}[t]
\centering
\begin{tabular}{@{}ccccc@{}}
\toprule
\multicolumn{2}{c}{Dataset}                         & \multicolumn{3}{c}{BGS}                                                     \\ \midrule
\multicolumn{2}{c}{Central}                         & \multicolumn{3}{c}{81.38$\pm$1.69}                                          \\ \midrule
\multicolumn{2}{c}{$K$}                             & 3                       & 5                       & 10                      \\ \midrule
\multicolumn{1}{c|}{\multirow{4}{*}{RE}}  & Local   & 67.36$\pm$3.66          & 67.04$\pm$1.79          & 64.54$\pm$1.65          \\
\multicolumn{1}{c|}{}                     & FedAvg  & 68.28$\pm$3.95          & {\ul 67.36$\pm$2.52}    & {\ul 65.04$\pm$1.26}    \\
\multicolumn{1}{c|}{}                     & FedProx & {\ul 68.28$\pm$2.96}    & 66.11$\pm$3.99          & 63.97$\pm$1.66          \\
\multicolumn{1}{c|}{}                     & FedHGN  & \textbf{69.20$\pm$3.22} & \textbf{67.52$\pm$2.58} & \textbf{65.73$\pm$1.47} \\ \midrule
\multicolumn{1}{c|}{\multirow{4}{*}{RET}} & Local   & 75.63$\pm$4.07          & 67.73$\pm$1.19          & {\ul 67.11$\pm$0.52}    \\
\multicolumn{1}{c|}{}                     & FedAvg  & 76.09$\pm$1.34          & 69.10$\pm$2.11          & 65.04$\pm$0.96          \\
\multicolumn{1}{c|}{}                     & FedProx & {\ul 76.32$\pm$2.13}    & \textbf{70.62$\pm$1.42} & 66.07$\pm$0.80          \\
\multicolumn{1}{c|}{}                     & FedHGN  & \textbf{78.93$\pm$2.23} & {\ul 69.79$\pm$2.07}    & \textbf{69.03$\pm$1.44} \\ \bottomrule
\end{tabular}
\caption{Experimental results (weighted average of accuracy \%) of node classification on the BGS dataset.}
\label{tab:node_classification_BGS}
\end{table}

\begin{table}[t]
\centering
\begin{tabular}{@{}ccclcc@{}}
\toprule
\multirow{2}{*}{SWD} & \multirow{2}{*}{CA} & \multicolumn{2}{c}{MUTAG (RE)}                       & \multicolumn{2}{c}{BGS (RE)}                      \\ \cmidrule(l){3-6} 
                     &                     & $K=5$                   & \multicolumn{1}{c}{$K=10$} & $K=5$                   & $K=10$                  \\ \midrule
B                    & \xmark              & 63.18$\pm$2.73          & 60.59$\pm$1.28             & 66.39$\pm$3.12          & 59.36$\pm$2.54          \\
C                    & \xmark              & {\ul 65.88$\pm$0.46}    & 61.26$\pm$1.33             & \textbf{67.78$\pm$2.09} & 64.11$\pm$2.40          \\
B+C                  & \xmark              & 65.29$\pm$0.93          & 62.83$\pm$3.13             & 66.53$\pm$2.93          & 63.76$\pm$0.61          \\
\xmark               & \xmark              & 65.06$\pm$0.34          & {\ul 65.47$\pm$0.74}       & 67.36$\pm$2.52          & {\ul 65.04$\pm$1.26}    \\ \cmidrule(r){1-3} \cmidrule(l){5-6} 
B                    & \cmark              & \textbf{67.87$\pm$0.83} & \textbf{65.59$\pm$0.86}    & {\ul 67.52$\pm$2.58}    & \textbf{65.73$\pm$1.47} \\ \bottomrule
\end{tabular}
\caption{Ablation study on the MUTAG and BGS datasets (RE splitting strategy). B: aggregating basis weights. C: aggregating schema-specific coefficients. The last row is our proposed FedHGN.}
\label{tab:ablation_full}
\end{table}

\end{document}